\documentclass[review]{elsarticle}

\usepackage{hyperref}
\usepackage{subfigure}
\usepackage{amsmath}
\usepackage{multirow}
\usepackage{booktabs}
\usepackage{xcolor}

\DeclareMathOperator*{\argmax}{arg\,max}
\usepackage{algorithm}
\usepackage{algpseudocode} 

\makeatletter
\def\ps@pprintTitle{%
  \let\@oddhead\@empty
  \let\@evenhead\@empty
  \let\@oddfoot\@empty
  \let\@evenfoot\@oddfoot
}
\makeatother
\begin{document}

\begin{frontmatter}

\title{Federated Learning of Molecular Properties with Graph Neural Networks in a Heterogeneous Setting}

\author[mymainaddress]{Wei Zhu}
\ead{wzhu15@ur.rochester.edu}

\author[mymainaddress]{Jiebo Luo}
\ead{jluo@cs.rochester.edu}
\address[mymainaddress]{Department of Computer Science, University of Rochester}
\address[mysecondaryaddress]{Department of Chemical Engineering, University of Rochester}

\author[mysecondaryaddress]{Andrew D. White$^*$}
\ead{andrew.white@rochester.edu}

\begin{abstract}
Chemistry research has both high material and computational costs to conduct experiments. Intuitions are interested in differing classes of molecules, creating heterogeneous data that cannot be easily joined by conventional methods. This work introduces federated heterogeneous molecular learning. Federated learning allows end-users to build a global model collaboratively while keeping their training data isolated. We first simulate a heterogeneous federated learning benchmark (FedChem) by jointly performing scaffold splitting and latent Dirichlet allocation on existing datasets. Our results on FedChem show that significant learning challenges arise when working with heterogeneous molecules across clients. We then propose a method to alleviate the problem: Federated Learning by Instance reweighTing (FLIT(+)). FLIT(+) can align the local training across clients. Experiments conducted on FedChem validate the advantages of this method. This work should enable a new type of collaboration for improving AI in chemistry that mitigates concerns about sharing valuable chemical data.
\end{abstract}
\begin{keyword}
federated learning \sep molecular property prediction \sep graph neural network
\end{keyword}

\end{frontmatter}


\section{Introduction}
There is an increasing trend to apply machine learning for molecule property prediction to avoid the expense of experiments or reduce the tremendous computational costs required for accurate quantum-chemical calculations. 
A large focus has been on applying graph neural networks to predicting molecular properties~\cite{wu2018moleculenet,schutt2018schnet,gilmer2017neural,Klicpera2020Directional,satorras2021n,yang2021predicting}. These works assume a central server that has access to all data. 
However, such a centralized learning scenario may not represent how institutions share chemical data. Due to intellectual property concerns and the intrinsic value of chemical data, it can be difficult for academic labs, national labs, and private institutions to share their molecule datasets.
\begin{figure}
\centering
\includegraphics[width=0.90\textwidth]{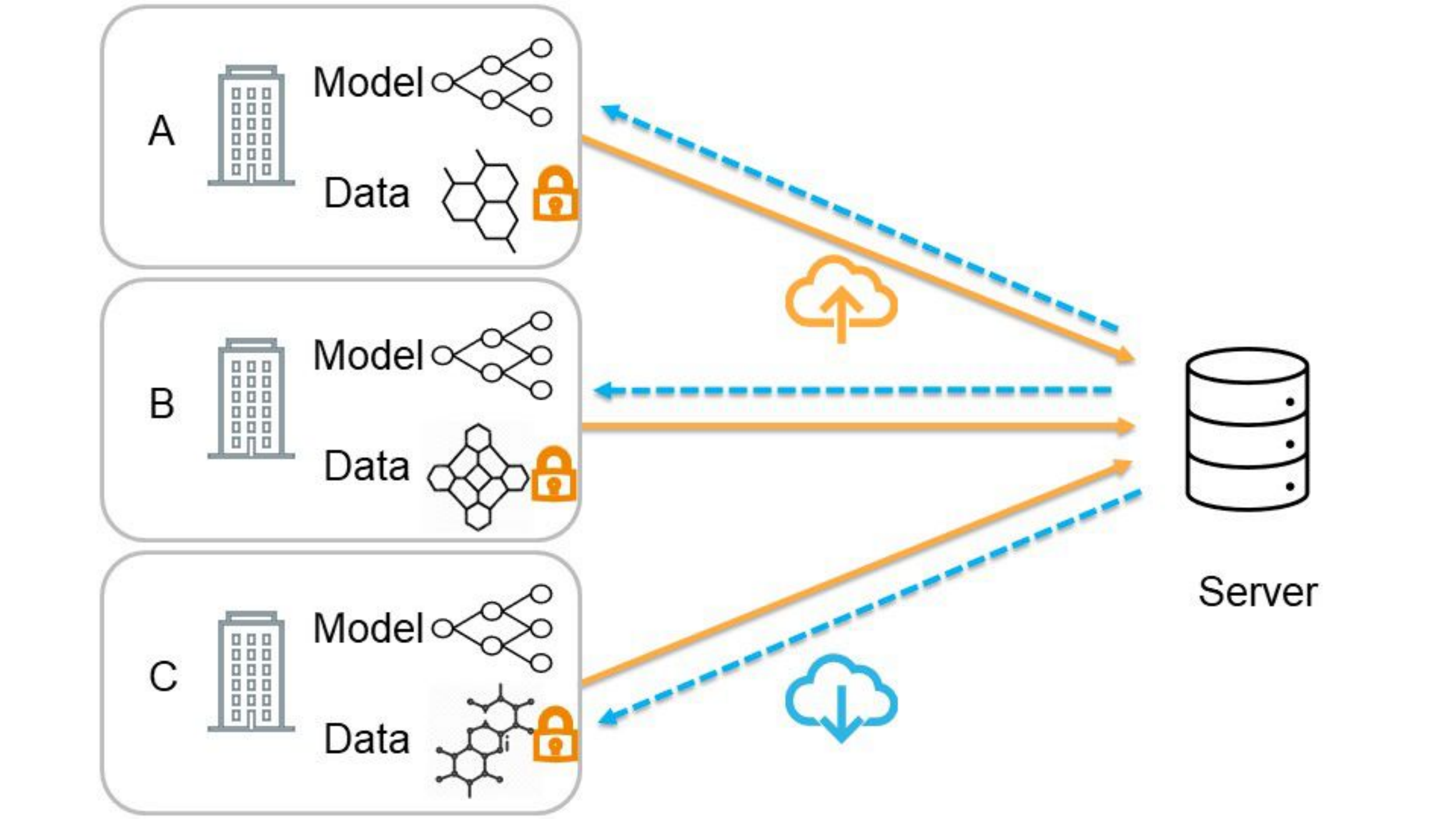}
\caption{We illustrate heterogeneous federated molecular learning where three institutions focus on different types of molecules. The server has no access to training data.}
\label{fig:demo}
\end{figure}

We propose federated learning to obtain a generalized global model without access to the private molecular data~\cite{mcmahan2017communication,he2021fedgraphnn}. For federated learning, local models are trained with their data on the client-side and then are aggregated for a global one on the server-side without seeing the data. One of the main concerns for federated molecular property prediction is the heterogeneously distributed client data since institutions focus on specific categories of molecules for their research interests. For example, institutions may wish to collaborate to construct an accurate model of pharmacokinetic clearance time of small molecules. Each institution studies specific drug-like molecules and their variants for their therapeutic targets. Each institution cannot share molecules, but it is beneficial to have a model for clearance time. Trained local models will heavily deviate from each other in this example, and it is thus sub-optimal to directly apply vanilla federated learning methods, e.g., Federated Average (FedAvg), to aggregate the heterogeneous local models~\cite{mcmahan2017communication}. Although several works are proposed to handle the heterogeneity problem~\cite{wang2020federated,xie2021federated}, a broader problem is the lack of heterogeneous federated molecular learning benchmarks to judge these methods for chemical data~\cite{wu2018moleculenet}.

This paper first proposes a federated heterogeneous molecular learning benchmark (FedChem). FedChem simulates the heterogeneous settings based on scaffold splitting~\cite{bemis1996properties} and Latent Dirichlet Allocation (LDA)~\cite{wang2020federated}. We first adopt scaffold splitting to split the molecules based on their two-dimensional structure, and molecules with similar structures are grouped accordingly~\cite{wu2018moleculenet}. Then, a heterogeneous setting is obtained by applying LDA on the scaffold subgroups, where LDA is a commonly used technique to simulate heterogeneous settings in conventional federated classification tasks~\cite{wang2020federated,he2020fedml}. We benchmark existing federated learning methods on the proposed heterogeneous suites FedChem and observer a remarkable performance degradation for the commonly used method FedAvg~\cite{mcmahan2017communication}. We then propose Federated Learning with Instance reweighTing (FLIT) to alleviate the heterogeneity problem by adapting focal loss for federated learning. The motivation of FLIT is that local models will be trained to overfit their data, which, however, do not share the same distribution as the global one. That is, the prediction of local models would be over-confident for certain types of molecules while with high uncertainty for others. FLIT can align the client training by adding weights to the uncertain cases by utilizing the local and received global models. As a result, the locally trained model will be more consistent with each other, and the federated learning performance can be eventually improved. We measure the uncertainty for training samples by the loss values and the prediction consistency among neighbored samples and develop two methods as FLIT and FLIT+ (FLIT(+) being the abbreviation for both). Our experiments on the proposed benchmark FedChem validate the advantages of FLIT(+) over existing federated learning methods.

Our  main contributions are summarized as follows:
\begin{enumerate}
    \item We propose a federated heterogeneous molecular learning benchmark based on MoleculeNet~\cite{wu2018moleculenet}, termed as FedChem. FedChem employs scaffold splitting and LDA to simulate the heterogeneous settings;
    \item We propose FLIT(+) algorithms to alleviate the heterogeneity problem. FLIT(+) can align the client training by putting more weights on uncertain samples;
    \item We conduct experiments to benchmark the proposed and existing federated learning methods on FedChem. Comprehensive experiments validate the effectiveness of the proposed methods. 
\end{enumerate}

\section{Related Work}
\subsection{Federated Learning}
 Federated learning was proposed by \cite{mcmahan2017communication} and has been applied in a wide range of fields including healthcare~\cite{chen2020fedhealth}, biometrics~\cite{aggarwal2021fedface}, natural images and videos~\cite{yang2019federated,deng2020fedvision}. As a popular method, Federated averaging (FedAvg) element-wisely aggregates the parameters of local models to obtain a global one ~\cite{mcmahan2017communication}. However, recent studies indicate that FedAvg may not handle the heterogeneity problem properly~\cite{wang2020federated,li2019convergence}. There are two categories of methods developed to alleviate the problem: improvements for server-side aggregation~\cite{wang2020federated,wang2020tackling,zhu2021data,reddi2020adaptive,lin2020ensemble,yurochkin2019bayesian,chen2020fedbe,li2019fedmd,mohri2019agnostic} and client-side regularization methods~\cite{karimireddy2020scaffold,dinh2020personalized,sahu2018convergence,chen2018federated,sarkar2020fed,reisizadeh2020robust}. 

Client-side methods can use the local training data and attract increasing attention. Our method also follows this line of research. Federated Proximal~(FedProx) regularizes the local learning with a proximal term to encourage the updated local model not to deviate significantly from the global model~\cite{sahu2018convergence}. A similar idea is adopted in personalized federated learning~\cite{dinh2020personalized}. SCAFFOLD adopts additional control variates to alleviate the gradient dissimilarity across different communication round~\cite{karimireddy2020scaffold}. Federated Model Distillation transfers the soft predictions of a shared dataset to reduce the communication cost and regularizes the local training with distillation loss~\cite{li2019fedmd}. Federated meta-learning incorporates MAML for local training to improve the generalization ability of local models~\cite{chen2018federated,fallah2020personalized}. Robust federated learning has been studied by several works~\cite{reisizadeh2020robust,mohri2019agnostic,li2019convergence}. FLRA adversarially conducts training on clients to make the model robust to affine distribution shifts~\cite{reisizadeh2020robust}. Most of the client-side federated learning methods add a regularization term to restrict the local training process so that the optimized local model would not deviate from the global one significantly~\cite{li2019fedmd,karimireddy2020scaffold,sahu2018convergence}. Consequently, the local models will be more consistent with each other, and the consistency could benefit the server-side aggregation. However, the regularization may also hinder the local optimization and lead to sub-optimal results for local training. Our method does not impose constraints on the local training and alternatively, we instance-wisely reweight the local training samples to align the local data distribution to the global one inspired by recent work~\cite{mukhoti2020calibrating,lin2017focal,sagawa2019distributionally,miyato2018virtual}. 

Heterogeneous federated learning is related to federated domain adaptation~(FDA)~\cite{DBLP:conf/iclr/PengHZS20,yao2022federated,song2020privacy}. FDA aims to improve the performance for specific target training domains, while general heterogeneous federated learning aims to improve the performance for all training data.

There are several works focusing on Federated Graph Neural Networks~\cite{xie2021federated,he2021fedgraphnn,lalitha2019peer,wang2020graphfl,wang2021fl,chen2020fede,peidecentralized} and federated molecular property prediction~\cite{xiong2020facing,MaLLWHXZ20}.  GraphFL applies MAML to improve the robustness of training~\cite{wang2020graphfl}.  The method in \cite{xie2021federated} alleviates the heterogeneity problem by group-wisely aggregating clients' models. However, existing work does not study federated molecular learning in heterogeneous settings where the clients' datasets are non-IID distributed in molecular structure and properties.

\subsection{Deep Molecular Property Prediction}
Graph neural network is commonly adopted for molecular learning ~\cite{wu2018moleculenet,schutt2018schnet,satorras2021n,hao2020asgn}. MPNN iteratively propagates the vertex features through message passing layers~\cite{gilmer2017neural}. SchNet adopts continuous-filter convolution to achieve E(3)-invariant molecular learning~\cite{schutt2018schnet}. DimeNet and DimeNet++ include directional information when training graph neural network for better performance~\cite{Klicpera2020Directional,klicpera2020fast}. Other works apply SO(3) equivariance message passing layer to predict the properties of molecular data~\cite{anderson2019cormorant,miller2020relevance}. A new structure is proposed by EQNN to efficiently achieve E(n) equivalent~\cite{satorras2021n}. We employ MPNN~\cite{gilmer2017neural} and SchNetThe~\cite{schutt2018schnet} for client-side training in the proposed federated molecular learning framework FedChem, and our framework can seamlessly integrate other models for client-side training, \textit{e.g.}, other graph network network~\cite{satorras2021n,Klicpera2020Directional,anderson2019cormorant} , sequence model~\cite{honda2019smiles,mayr2018large}, etc.

\section{Federated Heterogeneous Molecule Learning}

\subsection{Notations and Settings}
We first briefly describe Federated Heterogeneous Molecular Learning (FedChem). We assume that there are $L$ institutions that work on the same tasks with roughly different groups of molecules. That is, the data are distributed heterogeneously across institutions. Each institution develop a neural network for molecular property prediction~\cite{gilmer2017neural,schutt2018schnet,wu2018moleculenet}. The neural network trained on their data may suffer from poor generalization ability, and they thus intend to collaborate for a global model without sharing their data with the central server and other participants.

We propose to apply federated learning to obtain a global model for all participants without access to clients' data. Formally, we denote the overall dataset as $X=\lbrace X^l \rbrace_{l=1}^L$, where $X^l=(G^l, y^l)=\lbrace(g^l_i, y^l_i) \rbrace_{i=1}^{N_l}$ is the local dataset owned by the $l$-th institution/client which may not share the same distribution as the overall data. $g^l_i =(v^l_i, e^l_i)$ is the $i$-th molecule in graph representation with vertex as $v^l_i$, edge as $e^l_i$, groundtruth label as $y^l_i$. Ground truth could be either concrete values for regression tasks or categorical values for classification tasks. We utilize a local graph neural network $F^l$ to handle the data for the $l$-th client, and is implemented with Message Passing Neural Network (MPNN)~\cite{gilmer2017neural} or SchNet~\cite{schutt2018schnet}. To enable the clients to collaborate with each other, we have a central sever that receive and aggregate the uploaded local networks for a global one $F^g=FedAgg(\lbrace F^l \rbrace_{l=1}^L)$, where $F^g$ is the global model, and $FedAgg(\cdot)$ is the aggregation function, \textit{e.g.}, Federated Averaging~\cite{mcmahan2017communication}, Federated Optimzation~\cite{reddi2020adaptive}, Federated Distillation~\cite{seo2020federated}, FedDF~\cite{lin2020ensemble}, Federated Matched Averaging~\cite{wang2020federated}, etc. Note that the central server contains no training data, and also cannot access any local data.

FedChem simulates heterogeneous federated molecular learning with existing datasets, \textit{e.g.}, MoleculeNet~\cite{wu2018moleculenet}. Our method relies on scaffold splitting to group molecules based on their structure (graph). Molecules with similar structures are grouped into a scaffold subset. Scaffold splitting first groups the molecules into scaffold groups and then assign samples from each group to clients according to the unbalanced partition method Latent Dirichlet Allocation (LDA)~\cite{bemis1996properties}. We detail the approach to generating heterogeneous settings in the experimental section. 

Our method of generating a heterogeneous dataset is different from typical existing methods, which simulates label distribution shift~\cite{he2021fedgraphnn}. For example, \citet{karimireddy2020scaffold,wang2020federated} split samples based on class to each client, which makes the label distributions of local datasets on clients inconsistent with the global label distribution. In reality, institutions focus on molecules with similar structures via processes like lead optimization or hit finding\cite{jorgensen2009efficient}. Thus we typically see structurally heterogeneous molecules on the client-side (domain shift), while the label distributions among local clients can be similar. To simulate the structural heterogeneity with existing centralized datasets, we adopt scaffold splitting and do not rely on the ground-truth label. Intuitively, samples from different scaffold subsets are analogous to the samples from different domains for general machine learning tasks, and molecules~(images) within a scaffold subset~(domain) share similar structures~(style) but show different chemical properties~(ground-truth label). We illustrate the scaffold splitting to help readers better understand our heterogeneity simulation method. Moreover, it is non-trivial to generalize existing heterogeneous federated dataset simulation methods to regression and multi-label tasks, while our method can be easily adapted to any problems. We benchmark several existing federated learning methods on FedChem and observe that the heterogeneity problem brings significant challenges to federated molecular learning. 

\begin{algorithm}
	\caption{Federated Heterogeneous Molecule Learning (FedChem with FedAvg)}
	
	\begin{algorithmic}[1]
    	\Statex \textbf{Input:} \# clients $L$, \# local updates $T$, \# Comms round $C$
        \Statex \textbf{Output:} Global Model $F^g$
	    \State Server initialize a global model $F^g$ \Comment{Server init.}
		\While {Communication Round $< C$}
		\State Server broadcasts $F^g$ to clients
		\State $F^l \gets F^g$\Comment{Client init.}
		\For{$l:1$ to $L$ in parallel} \Comment{Client Update}
		\For {$t:1$ to $K$} \Comment{Update $F^l$ for $K$ steps}
		\State Sample a minibatch $\lbrace g^l_i, y^l_i \rbrace_{i=1}^B \sim X^l$
		\State Update local model $F^l$ by gradient descent
		\EndFor
		\State Client sends updated model $F^l$ to Server
		\EndFor
		\State Server gets $F^g \gets \sum_{l=1}^L \frac{|X^l|}{|X|} F^l$\Comment{Server Update}
		\EndWhile
	\end{algorithmic} 
	\label{alg:fedavg}
\end{algorithm} 
\subsection{Federated Learning with FedChem}
The basic training pipeline for FedChem is briefly introduced as follows: we first initialize a global model $F^g$ at server-side, and then for each federated learning communication round: 1). the server broadcasts global model $F^g$ to clients; 2). clients conduct training in parallel, and specifically, the $l$-th client is trained with its own data $X^l$ for an updated model as $F^l$; 3) the server collects updated local models from clients and then aggregate these models into a global one as $F^g=FedAgg(\lbrace F^i \rbrace_{i=1}^L)$. We iteratively perform steps 1-3 for $C$ communication rounds to obtain the final global model. We adopt FedAvg for server-side aggregation throughout the paper, but FedChem can be easily extended to involve other aggregation methods~\cite{reddi2020adaptive,wang2020federated}. We summarize the training procedure for federated learning with FedChem in Alg. \ref{alg:fedavg} by taking FedAvg as the aggregation method. Note that the server may select a subset of clients during each communication round for scalability. 

\subsection{Client-side Updates}


For completeness, we describe typical training steps to update the GNN model for client side training. We adopt MPNNs2s (MPNN set-to-set)~\cite{gilmer2017neural} and SchNet~\cite{schutt2018schnet} for molecule-level property prediction in our experiments, and other popular models (such as DimeNet~\cite{Klicpera2020Directional}, GIN~\cite{xu2018powerful}, GCN~\cite{kipf2016semi}, etc.) can also be unified in FedChem. 

Molecule-level GNN usually contains two phases: a message-passing phase and a readout phase~\cite{gilmer2017neural,he2021fedgraphnn}. Message-passing phase allows the vertex to propagate and collect information from their neighbors through the graph, and is usually composed of two steps as message generation and vertex update. Formally, given the $l$-th client model $F^l$ with $T$ message passing layers and a sampled graph $G^l$ (we omit the subscript for the sample, \textit{i.e.,} $G^l=G^l_i$), we define the message-passing function $M_t^l$ on the $i$-th vertex as~\cite{gilmer2017neural}
\begin{equation}
    m^l_{t+1,i} = M_t^l (  v^l_{t,i}, \lbrace v^l_{t,w}, e^l_{t,iw} \rbrace_{w\in N(i))}),
\end{equation}
 and the vertex update function $U_t^l$ as 
\begin{equation}
    v^l_{t+1,i} = U_t^l(v^l_{t,i}, m^l_{t+1,i}),
\end{equation}
where $v^l_{t,i}$ denotes the the representation of the $i$-th vertex in the $t$-th layer of $G^l$, $e^l_{t,iw}$ denotes the edge between the $i$-th and $w$-th vertex, $N(i)$ denotes the set of neighbors for vertex $i$ in graph $G^l$. $M_t^l(\cdot)$ generates the message  $m^l_{t+1,i}$ by aggregating the feature of $v^l_{t,i}$ and its neighbors, and also the edges between them. $U_t^l(\cdot)$ updates the $i$-th vertex by transforming the original features and the received message $m^l_{t+1,i}$. Different GNN models are implemented with different $M_t^l$ and $U_t^l$. For example, the message function of GCN is defined as $m^l_{t+1,i}=\sum_{w}{\hat{e}^l_{t,iw}FC(v^l_{t,w})}$ and $U_t^l=FC(m^l_{t,i})$~\cite{kipf2016semi}, where $FC$ is a linear layer and $\hat{e}$ is the Laplacian-regularized adjacency matrix. SchNet implements the message function $M_t^l$ with a continuous filter layer and  $U_t^l$ with a vertex(atom)-wise convolutional module~\cite{schutt2018schnet}. The message passing phase could aggregate and transform the vertex features for high level representations. 

After $T$ message passing layers, we adopt a readout function $R^l$ to aggregate the vertex representations for graph level representation as 
\begin{equation}
    {h}^l = R^l({v_{T,i}^l | i \in G^l}).
\end{equation}
$R^l$ should be permutation invariant and can be implemented with either a simple sum pooling or a learnable neural network. The graph-level representation $h^l$ is further used to obtain an estimation $\hat{y}^l=F^l(G^l)$ for the ground-truth molecular property $y^l$.

\begin{figure}
\centering
\includegraphics[width=0.8\textwidth]{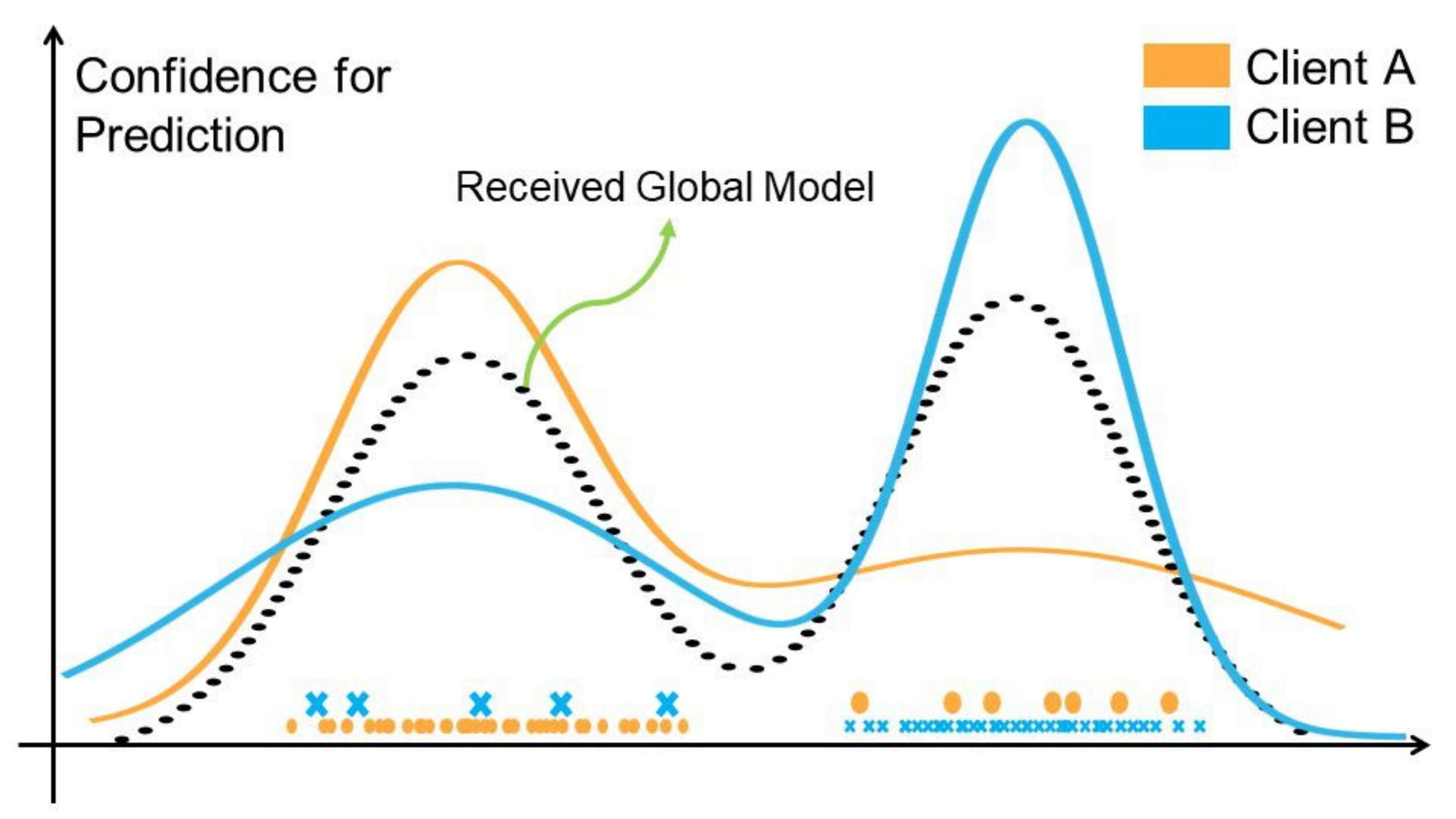}
\caption{Illustration for the motivation of FLIT. We assume two clients as $A$ and $B$, and the local data on these clients do not share the same distribution as the global one. Local models trained on biased local data will overfit the majority groups of data and underfit others. FLIT measures each sample's prediction confidence and puts more weight on the uncertain data. As a result, the local data distribution will be better aligned to the global one, and the trained local models will also be more consistent with each other.}
\label{fig:flit}
\end{figure}

\section{Federated Learning by Instance Reweighting FLIT(+)}
According to our experiments on the proposed heterogeneous federated learning benchmark FedChem,  heterogeneity brings significant difficulties to federated molecular learning. This section proposes a method to alleviate the heterogeneity problem, namely Federated Learning by Instance reweighTing (FLIT). FLIT adapts the formulation of focal loss for federated learning by involving a global model in local training objectives and can align the local training across clients by focusing on uncertain samples~\cite{sagawa2019distributionally,lin2017focal}. We illustrate the motivation of FLIT in Fig. \ref{fig:flit}.

Learning to reweight training samples is widely used in curriculum learning~\cite{zhou2020curriculum}, hard-sample Mining~\cite{lin2017focal}, domain generalization~\cite{sagawa2019distributionally,arjovsky2019invariant,pmlr-v139-krueger21a}, debiasing~\cite{nam2020learning}, model calibration~\cite{mukhoti2020calibrating}, adversarial defense~\cite{zhang2020geometry}, etc. Our method is closely related to Focal Loss~\cite{lin2017focal} and worst case optimization~\cite{sagawa2019distributionally}. Mukhoti \textit{et al.} point out that focal loss could make the objective value aligned with the prediction confidence~\cite{mukhoti2020calibrating}. GroupDRO improves the model generalization ability by assigning more weights for groups with the worst performance~\cite{sagawa2019distributionally}. 

FLIT relies on an instance reweighting framework to improve the federated molecular property prediction in a heterogeneous setting. The basic observation of FLIT is that, under the heterogeneous settings, the local model will be trained to overfit the small-scaled data at hand. Therefore, the local model will be over-confident for the majority groups of local training samples and may perform poorly and even worse than the received global model on the rare molecules at the client-side. As a result, the local models trained on different clients will significantly deviate from each other, and the inconsistency remarkably degrades the performance of the global model $F^g$, which is aggregated from the local models in a data-free manner~\cite{wang2020federated}. The suboptimal performance of FedAvg is wildly admitted by existing studies~\cite{wang2020federated,karimireddy2020scaffold}. FLIT puts more weight on samples with low prediction confidence by utilizing the local and global models to alleviate the problem. FLIT explores two different ways to define the prediction confidence, \textit{i.e.} the loss value~(FLIT) and also augmented with prediction consistency among the neighbors~(FLIT+). By focusing on the identified uncertain samples, FLIT(+) makes the local training more consistent across clients and eventually leads to better-federated learning performance. 

\begin{algorithm}
	\caption{FLIT(+) for $l$-th Client Updates} 
	\begin{algorithmic}[1]
		\Statex \textbf{Input:} $F^g$, $X^l=\lbrace(g^l_i, y^l_i) \rbrace_{i=1}^{N_l}$, $\gamma$ 
        \Statex \textbf{Output:} $F^l$
	    \State Save $\phi(g^l_i,F^g)$ Eq. (\ref{eq:flitphi})  or $\phi_{+}(g^l_i,F^g)$ Eq. (\ref{eq:flitplusphi})
	    \State $F^l\gets F^g$ \Comment{Init. $F^l$}
		\For {$t:1$ to $K$} \Comment{Train on the $l$-th Client}
		\State Sample a minibatch $\lbrace g^l_i, y^l_i \rbrace_{i=1}^B$
		\State calculate $\phi(g^l_i,F^l)$ by Eq. (\ref{eq:flitphi}) (or $\phi_{+}(g^l_i,F^l)$ by Eq. (\ref{eq:flitplusphi}))
		\State Obtain $\omega_{(+)}(x^l, F^l, F^g)$ by Eq. (\ref{eq:generalized_ir})
		\State $\omega(G^l_i, F^l, F^g) \gets \frac{ \omega(G^l_i, F^l, F^g)}{\bar{\omega}(G^l_i, F^l, F^g)}$ \Comment{Normalize $\omega_{(+)}$}
		\State Update $F^l$ by optimizing Eq. (\ref{eq:ir}) (or Eq.(\ref{eq:flitplusobj}))
		\State $\bar{\omega}_{(+)} \gets \beta \bar{\omega}_{(+)}+(1-\beta)\frac{1}{B}\sum_i {\omega_{(+)}}$ \Comment{Update moving average $\bar{\omega}_{(+)}$}
		\EndFor
	    \State Client sends updated model $F^l$ to Server
	\end{algorithmic} 
	\label{alg:flit}
\end{algorithm} 

\subsection{Federated Learning by Instance Reweighting}
 By jointly using the local model $F^l$ and global model $F^g$, FLIT reweights training samples to align the biased local data distribution to the global one. Eventually, the local models across clients will be well-aligned for better performance.

Given a molecule $x^l=(g^l, y^l)$ sampled from the dataset of the $l$-th client $X^l$, the original focal loss for binary classification tasks is defined as~\cite{lin2017focal}
\begin{equation} \label{eq:focal}
    \mathcal{L}_{focal}{(x^l)} = -(1-{\hat{y}}_t^l)^\gamma \log({\hat{y}}_t^l),
\end{equation}
where $\hat{y}_t^l$ is defined based on the prediction of molecule $\hat{y}^l=F^l(g^l)$ as
\begin{equation*}
\hat{y}_t^l = \begin{cases}
\hat{y}^l &\text{if $y^l=1$}\\
1-\hat{y}^l &\text{otherwise}.
\end{cases}
\end{equation*}
By substituting the binary cross entropy loss $\mathcal{L}(\hat{y}^l, y^l)=-\log(\hat{y}_t^l)$ into Eq. (\ref{eq:focal}), we have 
\begin{equation}
    \mathcal{L}_{focal}{(x^l)} = (1-\exp(-\mathcal{L}(\hat{y}^l, y^l)))^\gamma \mathcal{L}(\hat{y}^l, y^l).
\end{equation}
A generalized formulation for instance-reweighting can then be obtained as 
\begin{equation} \label{eq:ir}
        \mathcal{L}_{FLIT}{(x^l)} = (1-\exp(-\omega(x^l, F^l, F^g)))^\gamma \mathcal{L}(\hat{y}^l, y^l),
\end{equation}
where $\omega(x^l, F^l, F^g)$ is a non-negative function that indicates the uncertainty of training samples and is defined by jointly utilizing the local model $F^l$ and global model $F^g$ as 
\begin{equation} \label{eq:generalized_ir}
    \omega(x^l, F^l, F^g) = \phi(x^l, F^l) +  \max( \phi(x^l, F^l)- \phi(x^l, F^g),0),
\end{equation}
where $\phi(x,F)$ indicates the prediction uncertainty of $x$  with the model $F$. Eq. (\ref{eq:generalized_ir}) puts more weights on samples if the updated local model is less confident than the global model. We note $\omega(x^l, F^l, F^g)$ can take other types of formulation and we implement it with Eq. (\ref{eq:generalized_ir}) for simplicity. Moreover, for FLIT, we follow the focal loss and define $\phi(\cdot)$ as the loss value~\cite{lin2017focal}, \textit{i.e.}, 
\begin{equation}\label{eq:flitphi}
    \phi(x^l,F)=\mathcal{L}(\hat{y}^l, y^l).
\end{equation}
We substitute Eq. (\ref{eq:flitphi}) into Eq. (\ref{eq:generalized_ir}) and Eq. (\ref{eq:ir}), and the resulted method is termed as FLIT. Compared with the vanilla Focal loss, FLIT integrates the global model $F^g$ into the local training, which turns out to benefit the federated learning according to our experiments.
\subsection{FLIT+}
An alternative way to define $\phi(\cdot)$ for sample $x^l$ is the prediction discrepancy between the sample and its neighbors~\cite{wei2020theoretical}. Intuitively, the larger the discrepancy is, the less confident the model is for predicting the sample. To measure the prediction discrepancy for the neighborhoods, we aim to search for the data pairs with largest prediction discrepancy in the neighborhoods. Since directly searching for the exact neighbor is computationally expensive and is implausible with the local biased dataset, we alternatively adopts adversarial neighbor inspired by VAT (Virtual Adversarial Training)~\cite{miyato2018virtual}. Adversarial neighbors are similar to $x^l$ in terms of the input $g^l$ but has the most different prediction~\cite{wei2020theoretical}. Concretely, we measure the discrepancy by adversarial learning with a given model $F$ for $x^l$ as ~\cite{miyato2018virtual}
\begin{equation}
\begin{aligned}
  \Delta(x^l,F)=& D(F(g^l), F(g^l+ \xi r_{adv})) \\
  where \; r_{adv} &= \argmax_{r;\Vert r \Vert \leq \epsilon} D(F(g^l), F(g^l+ r)), 
\end{aligned}
\label{eq:vat}
\end{equation}
where $\epsilon=0.0001$ is a small positive value, $\xi=2.5$ is the step size, $D(\cdot)$ can be KL divergence for classification or Euclidean distance for regression~\cite{miyato2018virtual}. Eq. (\ref{eq:vat}) measures the discrepancy between predictions of the molecule with graph $g^l$ and its virtual adversarial neighbor $g^l+\xi r_{adv}$. Eq. (\ref{eq:vat}) generates a virtual adversarial neighbor $g^l+\xi r_{adv}$ that is similar to $g^l$ (since $\epsilon$ is small) but with most different prediction. We optimize $r$ on the positions for QM9 and vertex features for other datasets. We omit detail steps for optimizing Eq. (\ref{eq:vat}), and please refer to \cite{miyato2018virtual} for detail. We jointly use the loss value and the discrepancy defined in Eq. (\ref{eq:vat}), and obtain
\begin{equation} \label{eq:flitplusphi}
    \phi_{+}(x^l,F)=\mathcal{L}(\hat{y}^l, y^l)+\lambda\Delta(x^l, F)),
\end{equation}
where $\lambda$ is a hyperparameter. By substituting the formulation $\phi_{+}$ into Eq. (\ref{eq:generalized_ir}), we obtain $\omega_{+}(x^l, F^l, F^g)$ to measure the uncertainty of the training samples, and accordingly, we obtain FLIT+ by optimizing the objective as
\begin{equation} \label{eq:flitplusobj}
\begin{split}
    \mathcal{L}_{FLIT+}{(x^l)} 
    = (1-\exp(-\omega_{+}(x^l, F^l, F^g)))^\gamma (\mathcal{L}(\hat{y}^l, y^l) + \Delta(x^l,F^l)).
\end{split}
\end{equation}
Including $\Delta(x^l,F^l))$ in the training objective is essential to make the neighborhood prediction consistency a valid uncertainty measurement. Moreover, in experiments, we notice that federated learning can benefit from the virtual adversarial training alone \textit{i.e.} setting $\gamma=0$. This should be attributed to the fact that virtual adversarial training could improve the generalization ability of the local model and can be regarded as another way to align the local training implicitly. Detailed results and analysis can be found in the experimental section.

We use FLIT(+) to denote both FLIT and FLIT+. We summarize FLIT(+) for client update in Alg. \ref{alg:flit}.

\subsection{Implementation Details}
Since the scale of $\omega_{(+)}$ may vary significantly especially for regression tasks, it is not proper to directly applying Eq. (\ref{eq:ir}) and Eq. (\ref{eq:flitplusobj}) for general tasks. We propose to normalize the $\omega_{(+)}(\cdot)$ by its moving average as $\omega_{(+)} \gets \frac{ \omega_{(+)}}{\bar{\omega}_{(+)}}$, where
\begin{equation}
 \bar{\omega}_{(+)} \gets \beta \bar{\omega}_{(+)}+(1-\beta)\frac{1}{B}\sum_i {\omega}_{(+)}   
\end{equation}
 is the moving average and $B$ is the size of minibatch, $\beta$ is set as 0.8 in this paper. 

Moreover, we note that the prediction $F^g(g^l)$ and the discrepancy $\Delta(x^l,F^g)$ for the received global model only need to be calculated once per communication round and thus will not bring much computational cost.

\section{Experimental Procedures}

\subsection{Datasets}
We conducted experiments on a total of nine datasets retrieved from MoleculeNet~\cite{wu2018moleculenet} for molecular property prediction, including four regression datasets (FreeSolv, Lipophilicity, ESOL, and QM9) and five classification datasets (Tox21, SIDER, ClinTox, BBBP, and BACE). We follow the prediction tasks in \cite{wu2018moleculenet} and summarize the statistics for all datasets in Table \ref{tab:dataset}. 
\begin{table}[]
\centering
\begin{tabular}{@{}lcccc@{}}
\toprule
Dataset       & \#Compounds & \#tasks & task type & Metric  \\ \midrule
FreeSolv     & 642         & 1       & Reg.      & RMSE    \\
Lipophilicity & 4200        & 1       & Reg.      & RMSE    \\
ESOL          & 1128        & 1       & Reg.      & RMSE    \\
QM9           & 133885      & 12      & Reg.      & MAE     \\ \hline
Tox21         & 7831        & 12      & Cls.      & ROC-AUC \\
SIDER         & 1427        & 27      & Cls.      & ROC-AUC \\
ClinTox       & 1478        & 2       & Cls.      & ROC-AUC \\
BBBP          & 2039        & 1       & Cls.      & ROC-AUC \\
BACE          & 1213        & 1       & Cls.      & ROC-AUC \\ \bottomrule
\end{tabular}%
\caption{Statistics of datasets. Reg. and Cls. stand for regression and classification, respectively.}
\label{tab:dataset}
\end{table}
\begin{table*}[]
\centering
\resizebox{1\textwidth}{!}{%
\begin{tabular}{@{}lc|cc|ccc|cc|cc@{}}
\toprule
\multirow{2}{*}{Dataset}   &       & \multicolumn{2}{c}{Centeralized Training}        & \multicolumn{6}{c}{Federated Learning}                                      \\ 
                           & $\alpha$ & MolNet*                & FedChem*$_{\textrm{ours}}$ & FedAvg & FedProx &MOON & FedFocal$_{\textrm{ours}}$ & FedVAT$_{\textrm{ours}}$ & FLIT$_{\textrm{ours}}$ & FLIT+$_{\textrm{ours}}$ \\ \hline
         
\multirow{3}{*}{FreeSolv$\Downarrow$} & 0.1   & \multirow{3}{*}{1.40}  & \multirow{3}{*}{1.430}  & 1.771  & 1.693 &1.376   & 1.686       & 1.371     &    1.634           & \textbf{1.228}          \\
                           & 0.5   &                        &                         & 1.445  & 1.376  &1.423 & 1.322       & 1.299     &    1.366           & \textbf{1.127}          \\
                           & 1     &                        &                         & 1.223  & 1.216 &1.469  & 1.294       & 1.150     &     1.277          & \textbf{1.061}          \\ \hline
\multirow{3}{*}{Lipophilicity$\Downarrow$} & 0.1 & \multirow{3}{*}{0.655} & \multirow{3}{*}{0.6290} & \textbf{0.6361} & 0.6403 &0.6426& 0.6403 & 0.6556 &0.6563  & {0.6392} \\
                           & 0.5   &                        &                         & 0.6306 & 0.6365  &0.6339 & 0.6351     & 0.6333    &0.6368               & \textbf{0.6270}          \\
                           & 1     &                        &                         & 0.6505 & 0.6474   &0.6442 & 0.6461     & 0.6488    &  0.6443             & \textbf{0.6403}         \\ \hline
\multirow{3}{*}{ESOL$\Downarrow$}      & 0.1   & \multirow{3}{*}{0.97}  & \multirow{3}{*}{0.6570}  & 0.8016 & 0.7702&\textbf{0.7537}  & 0.8022      & 0.7776    & 0.7788         & {0.7642}         \\
                           & 0.5   &                        &                         & 0.7524 & 0.7382 &0.7258 & 0.7708      & 0.7243    &0.7426               & \textbf{0.7119}         \\
                           & 1     &                        &                         & 0.7056 & 0.6828 &0.6751 & 0.6822      & 0.7253    & \textbf{0.6705}             & 0.6998         \\ \hline
\multirow{3}{*}{QM9$\Downarrow$}       & 0.1   & \multirow{3}{*}{0.0479$^\spadesuit$}  & \multirow{3}{*}{0.0890$^\clubsuit$} & 0.5889 & 0.6036 &0.5817 & 0.6164      & 0.5606    & 0.5713              & \textbf{0.5356}         \\
                           & 0.5   &                        &                         & 0.5906 & 0.5751 &0.5707  & 0.6059      & 0.5656    &0.5658               &\textbf{0.5222}         \\
                           & 1     &                        &                         & 0.5786 & 0.5691 &0.5808  & 0.5822      & 0.5602    &0.5621               & \textbf{0.5282}         \\ \hline
\end{tabular}%
}
\caption{Performance for federated molecular regression. $\Downarrow, \Uparrow$ indicate if lower or higher numbers are better. $*$ denotes the results are obtained with centralized training. $^\spadesuit$ denotes the results are retrieved from~\cite{Klicpera2020Directional} with seperate SchNet for each task. $^\clubsuit$ denotes the results are obtained by a single multitask network. Smaller $\alpha$ of LDA generates more extreme heterogeneous scenario. FedFocal and FedVAT are proposed in this paper as the variants of FLIT(+). Best federated learning results are highlighted in bold.}
\label{tab:mainresults_reg}
\end{table*}
\begin{table*}[]
\centering
\resizebox{1\textwidth}{!}{%
\begin{tabular}{@{}lc|cc|ccc|cc|cc@{}}
\toprule
\multirow{2}{*}{Dataset}   &       & \multicolumn{2}{c}{Centeralized Training}        & \multicolumn{6}{c}{Federated Learning}                                      \\ 
                          & $\alpha$ & MolNet*                & FedChem*$_{\textrm{ours}}$ & FedAvg & FedProx &MOON & FedFocal$_{\textrm{ours}}$ & FedVAT$_{\textrm{ours}}$ & FLIT$_{\textrm{ours}}$ & FLIT+$_{\textrm{ours}}$ \\ \hline

\multirow{3}{*}{Tox21$\Uparrow$}     & 0.1   & \multirow{3}{*}{0.829} & \multirow{3}{*}{0.8182} & 0.7705 & 0.7732 &0.7331 & 0.7696      & 0.7733    &0.7711               & \textbf{0.7802}         \\
                          & 0.5   &                        &                         & 0.7811 & 0.7774 &0.7461 & 0.7812      &  0.7787   &  0.7825             & \textbf{0.7870}          \\
                          & 1     &                        &                         & 0.7770  & 0.7775 &0.7457 &\textbf{0.7881}      & 0.7706    &0.7748               & 0.7806         \\ \hline
\multirow{3}{*}{SIDER$\Uparrow$}     & 0.1   & \multirow{3}{*}{0.638} & \multirow{3}{*}{0.6260} & 0.6029 & \textbf{0.6056} &0.5885 & 0.6016      &0.6027           & 0.6035              & 0.6038         \\
                          & 0.5   &                        &                         & 0.6011 & 0.5931  &0.5966 & 0.6086      &  0.5981    &  0.6096             & \textbf{0.6146}         \\
                          & 1     &                        &                         & 0.6011 & 0.6023 &0.5901 & 0.6003      & 0.6053          &  0.6072             & \textbf{0.6174}         \\ \hline
\multirow{3}{*}{ClinTox$\Uparrow$}   & 0.1   & \multirow{3}{*}{0.832} & \multirow{3}{*}{0.8903} & 0.7491 & 0.7540 &\textbf{0.7892}  & \textbf{0.7789}      & 0.7581          &0.7761         & 0.7775         \\
                          & 0.5   &                        &                         & 0.7521 & 0.7423 &\textbf{0.7917} & 0.7770      & 0.7614    &\textbf{0.7888}              & 0.7852         \\
                          & 1     &                        &                         & 0.7784 & 0.7791 &0.8001 & \textbf{0.8036}      &  0.7743         &0.7849     & 0.7993         \\ \hline
\multirow{3}{*}{BBBP$\Uparrow$}      & 0.1   & \multirow{3}{*}{0.690} & \multirow{3}{*}{0.8674} & 0.8361 & 0.8610 &\textbf{0.8737}  & 0.8550      & \textbf{0.8673}          & 0.8666     & 0.8663         \\
                          & 0.5   &                        &                         & 0.8594 & \textbf{0.8879} &0.8865 & 0.8726      & 0.8641          &  0.8671       & 0.8774         \\
                          & 1     &                        &                         & 0.8453 & \textbf{0.8557}  &0.8487 & 0.8378      &  0.8386         & 0.8515      & 0.8515         \\ \hline
\multirow{3}{*}{BACE$\Uparrow$}      & 0.1   & \multirow{3}{*}{0.806} & \multirow{3}{*}{0.8834} & 0.8203 & 0.8328 &0.8373 & 0.8253      &  0.8166         &  0.8242             & \textbf{0.8467}         \\
                          & 0.5   &                        &                         & 0.8212 & 0.8398 &0.8285 & 0.8332      & 0.8417          &    0.8516           & \textbf{0.8667}         \\
                          & 1     &                        &                         & 0.8486 & 0.8408 &0.8561 & 0.8497      & \textbf{0.8578}          &   0.8497            & 0.8561         \\ \bottomrule
\end{tabular}%
}
\caption{Performance for federated molecular classification. $\Downarrow, \Uparrow$ indicate if lower or higher numbers are better. $*$ denotes the results are obtained with centralized training. Best federated learning results are highlighted in bold.}
\label{tab:mainresults}
\end{table*}

\subsection{Compared Methods}
To justify the proposed benchmark FedChem, we compare our results with MoleculeNet (MolNet) for centralized training~\cite{wu2018moleculenet}. To validate the effectiveness of FLIT(+), we compare FLIT(+) with Federated Averaging (FedAvg)~\cite{mcmahan2017communication}, Federated Proximal (FedProx)~\cite{sahu2018convergence} and MOON~\cite{li2021model}. Moreover, we also implement two variants of FLIT(+) as Federated Averaging with Focal loss for client training (FedFocal) and Federated Averaging with VAT for client training (FedVAT). We describe the compared methods as follows:

\begin{enumerate}
    \item Federated Averaging~(FedAvg)~\cite{mcmahan2017communication} simply element-wisely aggregates the local models to a global one;
    \item Federated Proximal~(FedProx)~\cite{sahu2018convergence} regularizes the local training to alleviate the heterogeneity problem; 
    \item MOON~\cite{li2021model} applies contrastive learning for federated learning to correct the local training;

    \item Federated Focal~(FedFocal) is proposed in this paper and is a variant of FLIT. FedFocal applies focal loss~Eq.(\ref{eq:focal}) to local training and adopts FedAvg for server update. FedFocal is proposed to validate the effectiveness of involving the global model into local training as FLIT;

    \item Federated VAT~(FedVAT) is also proposed in this paper and is a variant of FLIT+. FedVAT jointly optimizes~Eq. (\ref{eq:vat}) and original training loss for client training and adopts FedAvg for server update. Compared with FLIT+, FedVAT does not use instance reweighting training strategy; 

    \item Federated Learning by Instance reweighTing~(FLIT) is proposed in this paper and is described in Algorithm \ref{alg:flit}. 

    \item FLIT+ is proposed in this paper. Compared with FLIT, FLIT+ jointly uses loss values and the discrepancy between nearby samples to measure the uncertainty of samples as described in~Eq. \ref{eq:flitplusphi}, and adopts~Eq. \ref{eq:flitplusobj} as the learning objective. 

\end{enumerate}
We perform grid search on the excluded validation set for hyperparameter tuning and model selection. For FedProx, we search the hyperparameter $\mu$ from $[0.001,0.01,0.1,1,10]$. For MOON, we search the hyperparameter from $[0.1,1,5,10]$. We search $\gamma$ used for instance reweighting for FLIT(+) and FedFocal from $[0.5,1,2]$, and search $\lambda$ from $[0.01,0.1,1]$ for FLIT+. FedVAT adopts a hyperparameter  to balance VAT loss and primary loss which is searched from $[0.01,0.1,1]$. We report results on the testing set by the model with the best performance on the  validation set. 

\subsection{Main Results}
The experimental results on regression and classification datasets are shown in Table \ref{tab:mainresults_reg} and Table \ref{tab:mainresults} respectively. We draw several points according to the results. First, comparing our centralized training results (denoted as FedChem) with MolNet~\cite{wu2018moleculenet}, we obtain competitive results by using MPNNs2s~\cite{gilmer2017neural} and SchNet~\cite{schutt2018schnet}. Specifically, we obtain a significant performance gain by adopting SchNet for QM9 dataset~\cite{schutt2018schnet}. Second, comparing the performance of FedAvg with different $\alpha$ for each dataset, we can conclude that the heterogeneity settings introduced by FedChem indeed lead to performance degradation for 7 out of 9 datasets (i.e., FreeSolv, ESOL, QM9, Tox21, ClinTox, BBBP, and BACE). FedAvg shows stable performance for Lipophilicity and SIDER. The reason may be that we do not consider the relation between scaffold subgroups in our current settings, and the resulted clients' datasets are rather homogeneous. Third, we observe a significant performance gain for most datasets by comparing heterogeneous federated learning methods with FedAvg. For example, the proposed FLIP+ achieves a 0.543 improvement with $\alpha=0.1$ and 0.162 improvement with $\alpha=1$ for FreeSolv. The results suggest the necessity to mitigate the heterogeneity when conducting federated learning and validate the effectiveness of the proposed FLIT(+). However, we also observe that the performance improvements of our methods are rather marginal for several datasets. The reasons may be attributed to the fact that our current scaffold splitting may not lead to heterogeneous datasets. We will continue our work for a better method to simulate the heterogeneity problem for federated molecular property prediction.   

Moreover, the proposed instance-reweighting methods (FedFocal, FLIT, and FLIT+) outperform the regularization-based methods FedProx and MOON. The proposed FLIT additionally utilizes the global model and performs better than its counterpart FedFocal. For example, FLIT improves FedFocal from 0.8022 to 0.7788 with $\alpha=0.1$ and from 0.7708 to 0.7426 with $\alpha=0.5$ for ESOL. Lastly, FLIT+ further improves the performance of FLIT by measuring the uncertainty with loss values and discrepancy between neighbors. We also observe that FedVAT can benefit federated learning by encouraging locality smoothness for better generalization performance. By incorporating VAT~\cite{miyato2018virtual} into the FLIT framework, FLIT+ achieves the best overall performance. FLIT(+) has more consistent results across different settings of $\alpha$ compared with its counterparts, indicating the effectiveness of FLIT+ for dealing with heterogeneity problems.
\begin{figure*}
\centering
\subfigure[ESOL$\Downarrow$]{
\includegraphics[width=0.30\textwidth]{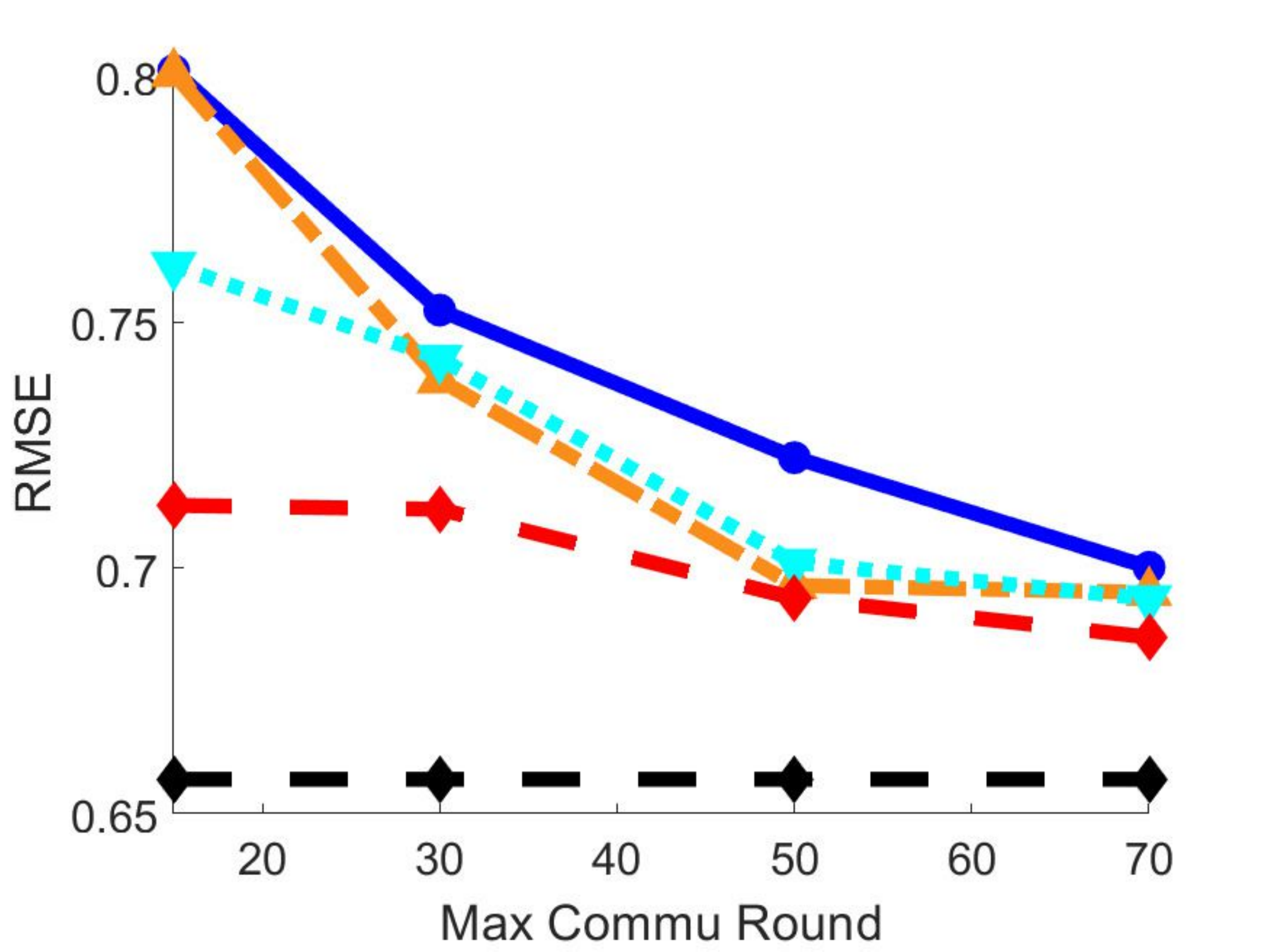}
}
\subfigure[ClinTox$\Uparrow$]{
\includegraphics[width=0.30\textwidth]{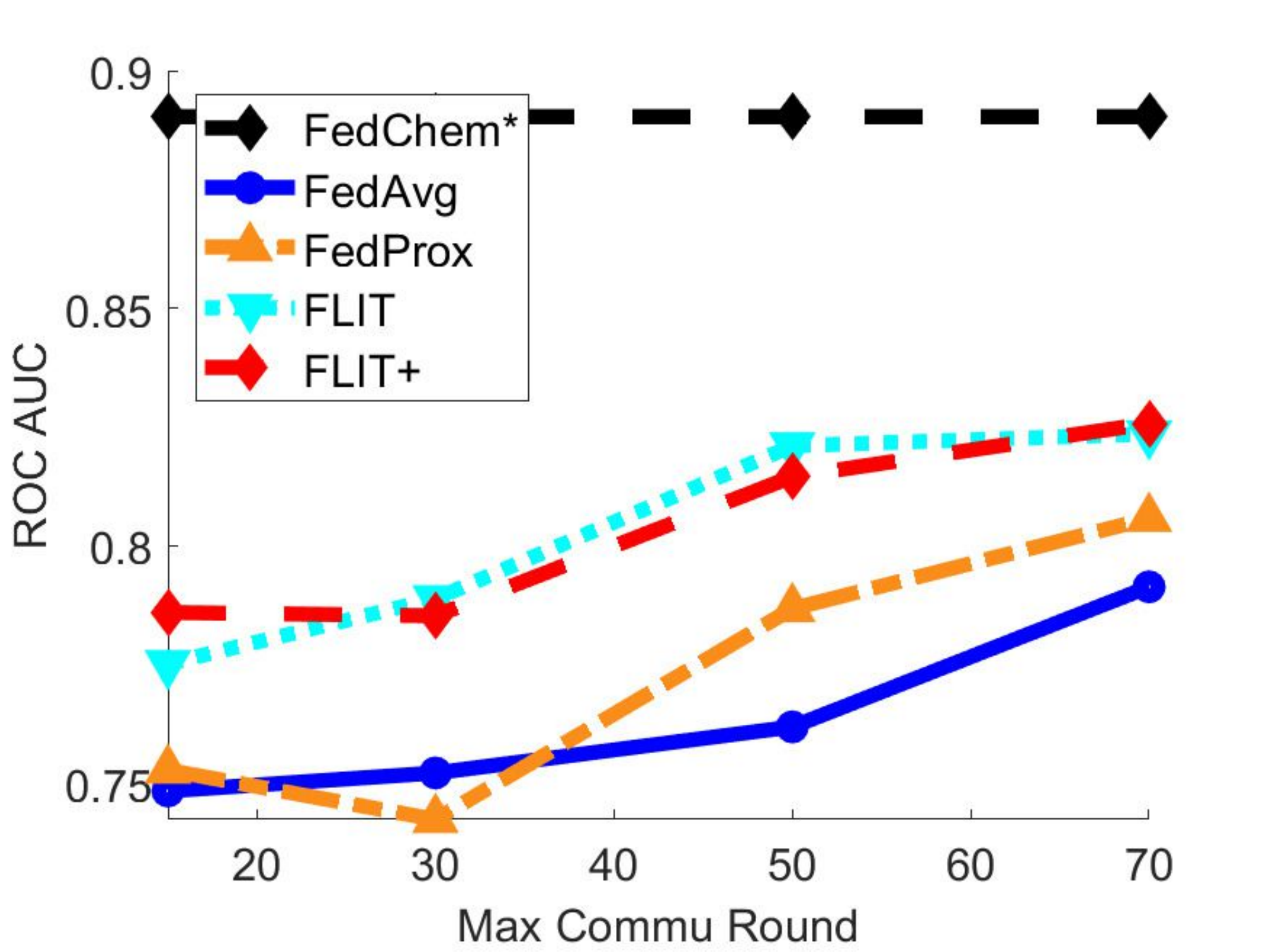}
}
\subfigure[BACE$\Uparrow$]{
\includegraphics[width=0.30\textwidth]{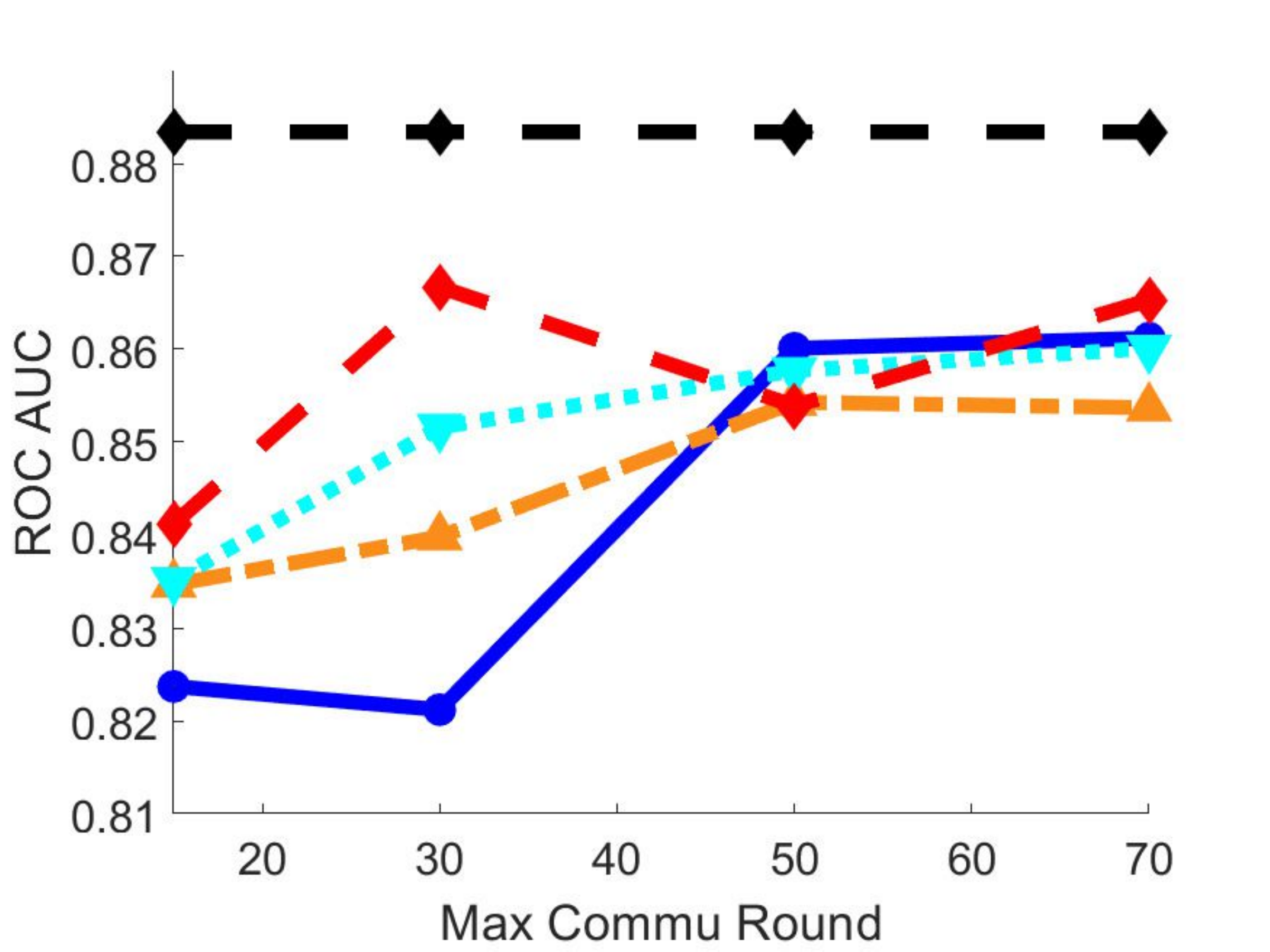}
}
\caption{Performance of baseline and our methods with varying communication rounds. $*$ denotes that the results are obtained with  centralized training. We find our method has a strong advantage with a few communication rounds.}
\label{fig:numround}
\end{figure*}

\subsection{Sensitivity Analysis for Federated Learning}
\begin{figure*}
\centering
\subfigure[ESOL$\Downarrow$]{
\includegraphics[width=0.30\textwidth]{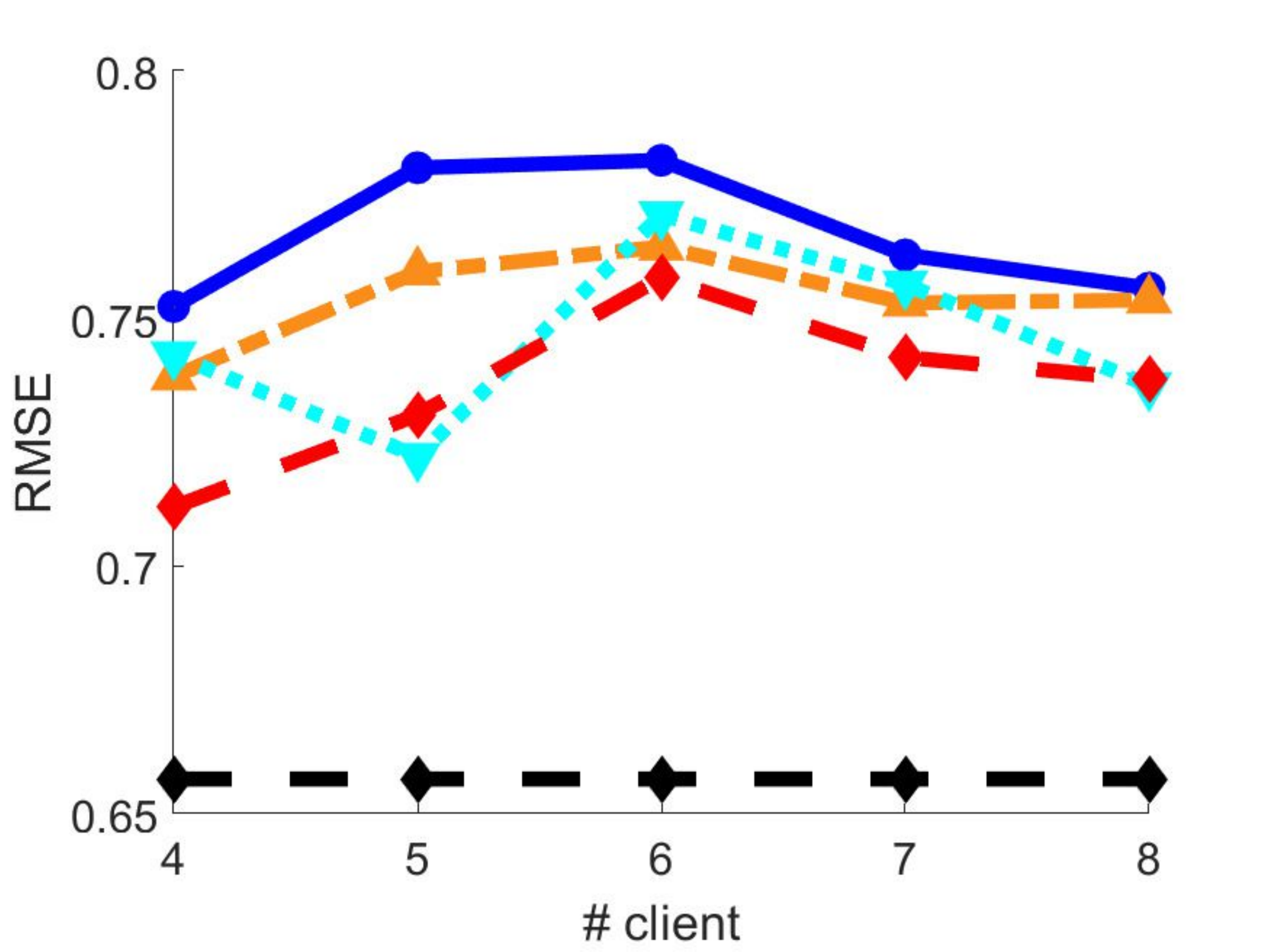}
}
\subfigure[ClinTox$\Uparrow$]{
\includegraphics[width=0.30\textwidth]{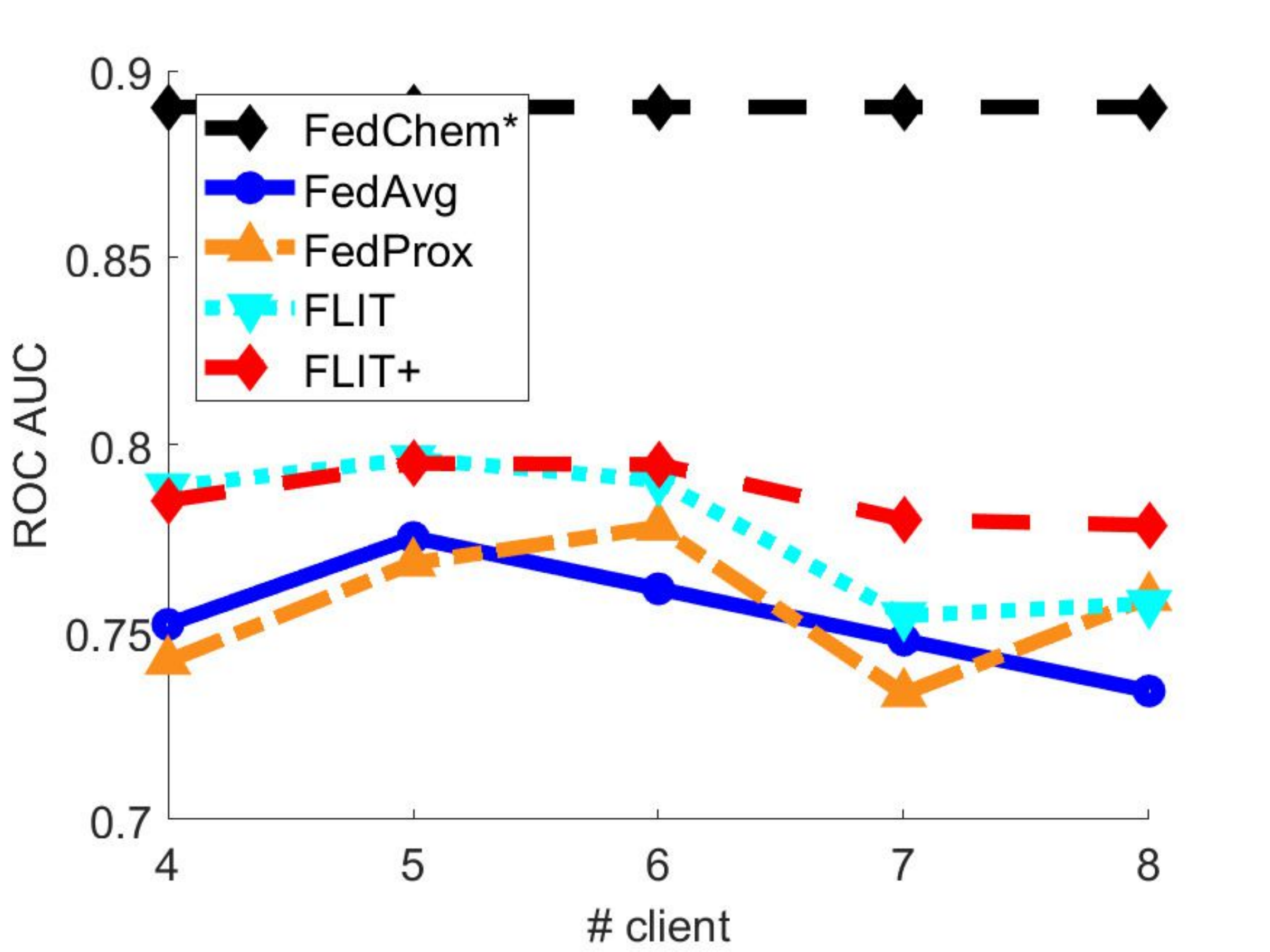}
}
\subfigure[BACE$\Uparrow$]{
\includegraphics[width=0.30\textwidth]{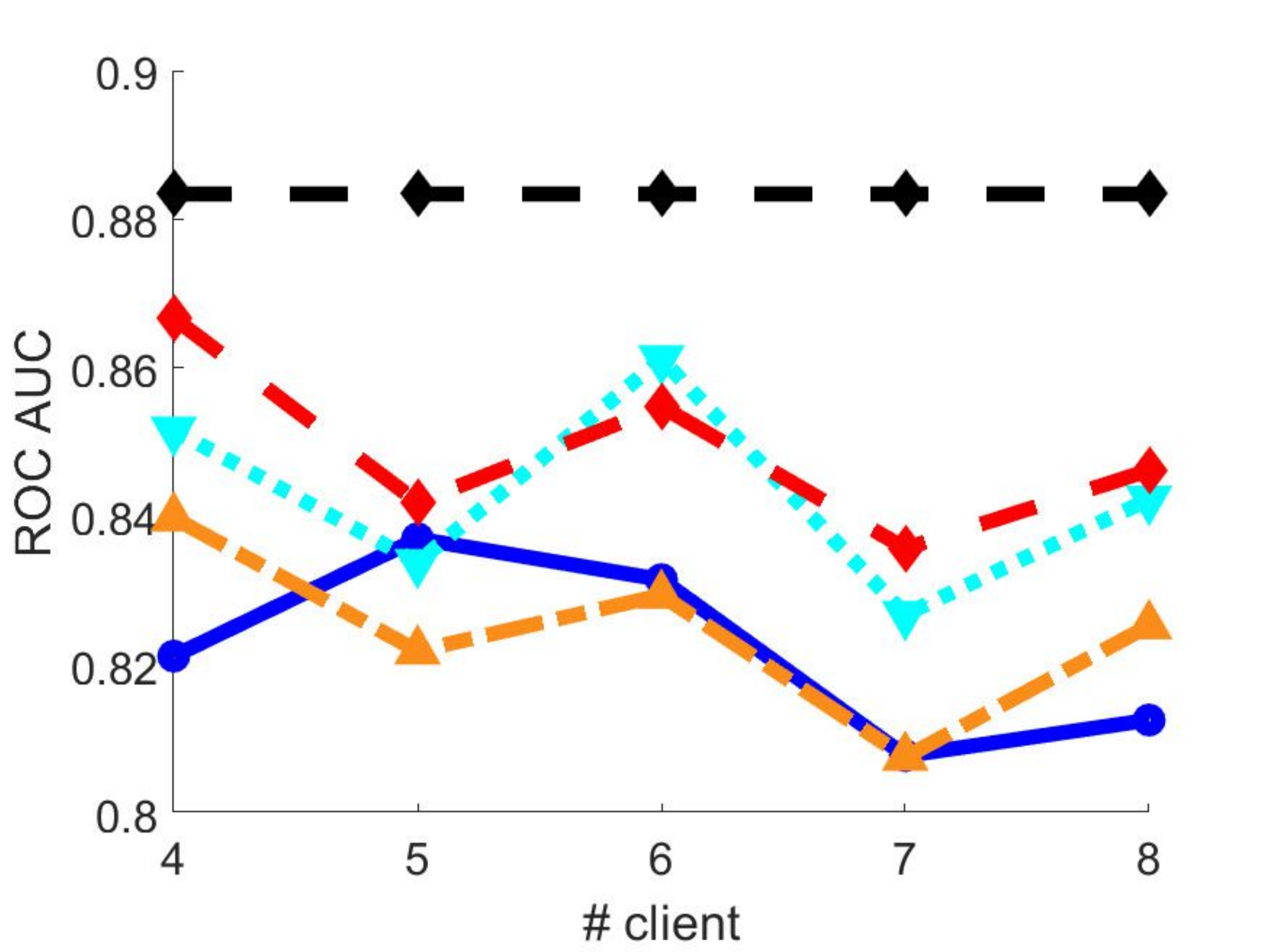}
}
\caption{Performance of baseline and our methods with different number of clients. See Figure~\ref{fig:numround} for color legend. The small scale local training data reduce federated learning performance for all methods. }
\label{fig:numclient}
\end{figure*}

This section studies the influence of the number of clients and communication rounds on the federated learning performance. For simplicity, we conduct experiments on ESOL, ClinTox, and BACE. The results of the different number of maximum communication rounds are shown in Fig. \ref{fig:numclient}. We vary the maximum communication round from $\lbrace 15,30,50 \rbrace$ while fixing the total local steps. We find that increasing the frequency of communication can benefit federated learning, although it also leads to increased transfer costs. The performance with different numbers of clients is shown in Fig. \ref{fig:numclient}. We vary the number of clients within $\lbrace 4,5,6 \rbrace$ since a large number of clients would lead to over small local datasets, which is not plausible for valid training. We find that the performance of federated learning usually decreases (ESOL and BACE) or is stable (ClinTox) as the client number increases. This indicates that small-scale local training data degrade the federated learning performance. 


\section{Settings for Heterogeneous FedChem}
For all datasets except QM9, we first randomly split the dataset into 80$\%$ for training, 10$\%$ for validation, and 10$\%$ for testing following~\cite{wu2018moleculenet}. QM9 is partitioned into 110,000 samples for training, 10,000 samples for validation, and the remaining for testing following~\cite{schutt2018schnet}. To simulate the heterogeneous settings for federated learning, we first perform scaffold splitting~\cite{bemis1996properties} to partition the training data into subgroups. Then, we assign the molecules of each subgroup to clients by Latent Dirichlet Allocation (LDA)~\cite{he2020fedml,wang2020federated}. We control the degree of heterogeneity by tuning $\alpha$ for LDA~\cite{wang2020federated}. Smaller $\alpha$ leads to more severe heterogeneity and we vary $\alpha$ from $\lbrace 0.1, 0.5, 1 \rbrace$. Moreover, we deliberately balance the number of molecules for each client following~\cite{he2020fedml} to control for the effect of the example number on performance~\cite{wang2020tackling}. 

As for federated learning settings, we set the default communication rounds $C$ to 30 and the default number of clients to four for all datasets except for QM9, which is set to eight. 

For client training, we set the batch size to 64 and use Adam~\cite{kingma2014adam} with a learning rate of $1\times10^{-4}$ and a weight decay of $1\times10^{-5}$.
For all datasets except QM9, we simulate four clients and train the local model for 10,000 local steps. QM9 has eight clients, and we train the model for 100,000 local steps. We conduct federated learning with the FedML framework~\cite{he2020fedml}.

For all datasets except QM9, we use MPNNs2s implemented by Deep Graph Library~\cite{wang2019deep}.  MPNNs2s has 3 message passing layers and 3 set2sets layers. The hidden features of the edge is 16, and the output feature of the vertex is 64. We perform three set2set steps. For QM9, we adopt SchNet with six interaction layers~\cite{schutt2018schnet} and implement SchNet by PyTorch-Geometric~\cite{FeydLenssend2019}. The number of hidden channels and filters of  SchNet is 128, and the number of Gaussian is set as 50 for continuous filter layers. We implement a multitask network for datasets with multi-objectives. We run experiments on a server with 8 NVIDIA RTX 2080 Ti Graphics Cards.

\section{Conclusions}
Chemistry can be a challenging domain for deep learning because of the computational and material cost per training example. For example, each row in the Tox21 dataset costs about \$50--\$300 million USD \cite{dimasi2016innovation}. Therefore, contributing data to a public dataset may be impossible for institutions due to the intrinsic value of the data. Federated learning is a way to build global models while preventing the dissemination of chemical data. We propose a benchmark called FedChem for heterogeneous chemical data, which mimics how chemical data distributes among institutions. FedChem is composed of regression and classification learning scenarios from the existing MoleculeNet dataset and utilizes scaffold splitting and LDA (Latent Dirichlet Allocation) to assign molecules with different structures to different clients. FedChem can be tuned to generate scenarios with different degrees of heterogeneity. Given that existing federated learning methods perform poorly on FedChem, we propose an instance re-weighting framework called Federated Learning with Instance reweighTing (FLIT(+)), inspired by focal loss to align the training process across clients. We show that FLIT(+) is robust to different tasks and datasets with extensive experiments. One possible future direction is to develop personalized federated learning for FedChem~\cite{xiong2020facing}. Moreover, since our current heterogeneous simulation method may not lead to severe structural heterogeneity problems in some cases, we will explore other approaches for more heterogeneous settings.

\section{Acknowledgements}
Research reported in this work was supported by
the National Institute of General Medical Sciences of
the National Institutes of Health under award number
R35GM137966.


\bibliographystyle{model5-names}
\bibliography{fed}

\end{document}